\documentclass[10pt,twocolumn,letterpaper]{article}

\usepackage[pagenumbers]{wacv} 

\definecolor{wacvblue}{rgb}{0.21,0.49,0.74}
\usepackage[pagebackref,breaklinks,colorlinks,allcolors=wacvblue]{hyperref}

\def\wacvPaperID{302} 
\def\confName{WACV}
\def\confYear{2027}

\title{PoseBench: Probing Pose Awareness in Visual Foundation Models}

\author{Most Nilufa Yeasmin\\
South Dakota State University\\
Brookings, SD, USA\\
{\tt\small MostNilufa.Yeasmin@jacks.sdstate.edu}
\and
Draix Wyatt\\
South Dakota State University\\
Brookings, SD, USA\\
{\tt\small Draix.Wyatt@jacks.sdstate.edu}
\and
Ruyi Lian\\
South Dakota State University\\
Brookings, SD, USA\\
{\tt\small Ruyi.Lian@sdstate.edu}
}

\begin{document}
\maketitle
   \begin{abstract}
Pose awareness characterizes the extent to which a learned feature space aligns with the underlying object-pose manifold, such that observations with similar poses exhibit greater feature similarity than those with increasingly different poses.
While such structure is desirable for pose-sensitive 3D vision tasks, visual foundation models pretrained on individual 2D images typically lack explicit 3D or pose supervision.
This raises a fundamental question: Do 2D visual foundation models exhibit pose awareness after large-scale pretraining?
To answer this question, we introduce PoseBench, a systematic benchmark to probe pose awareness in pretrained visual features. 
Our evaluation follows a progressive design: held-out rendered views probe intrinsic pose-aware structure under controlled conditions, while real-image queries evaluate how robustly this structure remains accessible under real-world variations. We further introduce pose-graph-based refinement as a complementary probe of compatibility between feature neighborhoods and the underlying pose manifold. 
Across 31 publicly available checkpoints from 10 pretraining families, PoseBench reveals that pose awareness is an implicit but non-uniform property of 2D visual foundation models, and consistent gains from pose-graph refinement suggest that feature spaces retain structure compatible with the underlying pose manifold. PoseBench therefore provides a systematic basis for understanding and selecting pretrained features for pose-sensitive 3D vision tasks.

\end{abstract}    
   \section{Introduction}
\label{sec:intro}

\begin{figure}[t]
  \centering
  \includegraphics[width=0.9\linewidth]{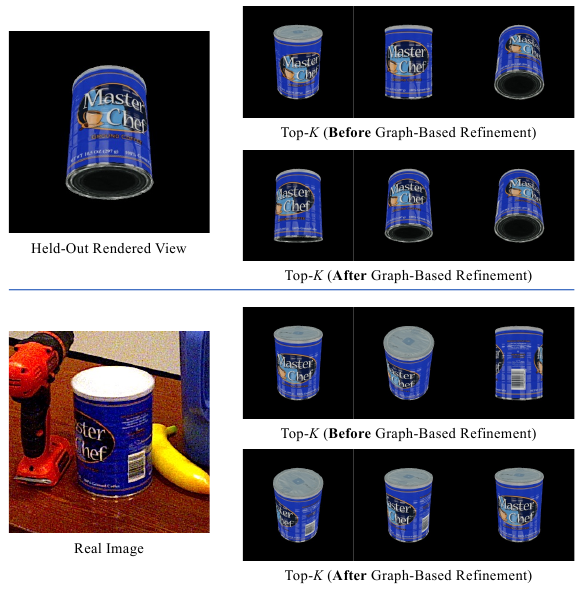}
  \caption{Illustrative overview of PoseBench. Given a query image (held-out template or real image), we assess pose awareness of foundation model features via retrieving top-$K$ template views with or without graph refinement. 
  Retrieval examples are shown for illustration only.}
  \label{fig:motivation}
\end{figure}

\begin{figure*}[t]
  \centering
  \includegraphics[width=0.95\textwidth]{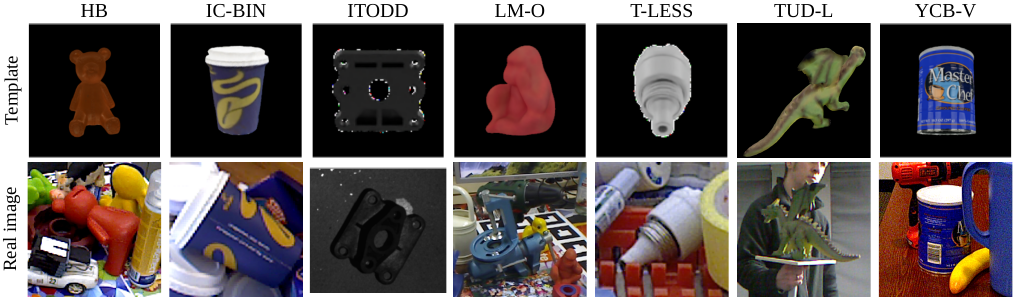}
  \caption{PoseBench builds on 7 BOP datasets~\cite{hodan2018bop} with
  substantial visual diversity, shown here as one representative
  object per dataset: a rendered template (top) and a real-world
  photograph of the same object (bottom). The benchmark ranges from
  toy figures on calibration boards (HB, TUD-L) to cluttered
  bin-picking scenes (IC-BIN), texture-less industrial parts
  (T-LESS), heavily occluded cluttered scenes (LM-O), grayscale
  industrial sensor captures (ITODD), and household objects (YCB-V).}
  \label{fig:overview}
\end{figure*}

2D visual
foundation models~\cite{caron2021emerging, oquab2023dinov2, he2022mae,
radford2021clip, dosovitskiy2021vit} 
have become the default backbones for 
an enormous range of downstream tasks.
However, almost none of their pretraining objectives, such as self-distillation,
masked reconstruction, vision--language contrastive learning, or
supervised classification, explicitly supervise object pose or 3D geometry. Instead, many objectives encourage invariance to pose-related transformations. 
This motivates a question important to pose-sensitive applications: do visual foundation models exhibit \emph{pose awareness} after large-scale pretraining? 

In this work, we propose \emph{PoseBench}, a systematic benchmark
for probing pose awareness in pretrained visual features (Figure~\ref{fig:motivation}). 
Our benchmark builds upon the seven core datasets of the BOP Benchmark~\cite{hodan2018bop}, 
as shown in Figure~\ref{fig:overview}.
Following common practice in modern pose pipelines~\cite{liu2026survey}, we crop and rescale both rendered templates and real images to an object-centric square format.
This normalization yields objects at comparable scales and reduces variation due to image-plane translation. 
We therefore focus our analysis of pose awareness on object rotation, treating translation as a nuisance factor to which the representation should ideally remain invariant.
PoseBench probes pose awareness with a progressive design: held-out rendered views first
probe intrinsic structure under controlled conditions; real-image
queries against the same gallery then test how robustly that
structure survives the sim-to-real domain gap, including occlusion
and illumination variation. We further introduce
pose-graph-based refinement, which propagates raw similarities
across known relative poses between templates, as a complementary
probe. 

We evaluate 31
checkpoints spanning 10 pretraining families (21M to
1.1B parameters) under this protocol.
The network depth carrying the strongest pose-aware signal consistently lies at an intermediate layer under controlled conditions, while shifts toward deeper layers as appearance variation is introduced.
Model-size scaling is non-monotonic: DINOv3-Small~\cite{simeoni2025dinov3} (21M
parameters) outperforms DINOv2-Giant~\cite{oquab2023dinov2} (1.1B parameters) on real
images.
Robustness to real imagery
also varies substantially by family. Finally, pose-graph-based
refinement improves every metric in every occlusion regime, including
heavy occlusion.
Together, these results establish pose awareness as an implicit but
non-uniform property of large-scale 2D visual representations.

In summary, our contributions are:
\begin{itemize}
  \item We propose a symmetry-aware formalization of
  pose-awareness, which isolates a representation's intrinsic pose-aware
  structure from any downstream solver.
  \item We present PoseBench, the first systematic analysis of pose-aware representations across 31 checkpoints from 10 model families.
  \item We demonstrate that pose awareness broadly presents in 2D visual foundation model features, and it
  scales non-monotonically with model size.
  \item We develop graph refinement as a complementary probe, showing that the feature space retains overall pose structure and local noise can be mitigated by refinement.
  \item We quantify sim-to-real domain gap and occlusion robustness and reveal cross-family complementarity.
\end{itemize}

   \section{Related Work}
\label{sec:related}

\noindent\textbf{Visual foundation models.}
Large-scale pretraining has produced ViT backbones~\cite{dosovitskiy2021vit} optimized for
a diverse range of objectives.
\emph{Self-distillation} methods train a student network to match the
output of a teacher network 
across different augmented crops of the same image.
DINO~\cite{caron2021emerging} established this recipe and is known to
produce features with strong emergent correspondence and part-level
structure even without any task-specific probe; DINOv2~\cite{oquab2023dinov2}
scales it to a much larger, curated dataset and adds register tokens
in a later variant~\cite{darcet2024vitneedreg} to suppress high-norm
artifact tokens that otherwise degrade dense features, and
DINOv3~\cite{simeoni2025dinov3} scales the same recipe further still.
\emph{Masked Image Modeling} (MIM) methods instead mask a large
fraction of an image's patches and train the network to reconstruct
what is missing, which forces the encoder to retain the fine-grained
spatial and structural detail needed for pixel-accurate
reconstruction: MAE~\cite{he2022mae} reconstructs raw pixel values
directly, while BEiT~\cite{bao2022beit} instead predicts discrete
visual tokens from a pretrained tokenizer, trading some of that pixel
fidelity for a more compressed, semantically-abstracted target.
\emph{Vision--language
contrastive} methods such as CLIP~\cite{radford2021clip} instead align
a whole image with its paired text caption in a shared
embedding space, trained on hundreds of millions of web image-text
pairs.
EVA-02~\cite{fang2023eva02} is a hybrid of the two ideas
above, using masked-image-modeling to reconstruct the features
of a frozen CLIP teacher rather than pixels or tokens, aiming for
CLIP's semantic richness with MIM's spatial grounding. Finally,
\emph{supervised classification} remains a strong baseline objective.
The original ViT~\cite{dosovitskiy2021vit} trains a plain
transformer end-to-end on labeled ImageNet~\cite{deng2009imagenet}, and
DeiT~III~\cite{touvron2022deit3} refines the training recipe
(augmentation, regularization, and optimization).

\noindent\textbf{Benchmarking foundation model features.}
A growing body of work asks what pretrained representations capture
beyond their training objective, typically through frozen probing
rather than fine-tuning.
Kornblith~\etal~\cite{kornblith2019better}
established this protocol for classification transfer, probing
representations layer-by-layer to test whether ImageNet accuracy
predicts downstream usefulness; in a different vein,
Amir~\etal~\cite{amir2021deep} showed that DINO's raw patch
descriptors, used entirely as-is with no probe or fine-tuning at all,
already support dense correspondence, co-segmentation, and part
discovery. Battle of the Backbones~\cite{goldblum2023battle}
scales this style of inquiry to dozens of pretrained models across
classification, detection, and segmentation, finding that supervision
type and training recipe matter more than scale.
Walmer~\etal~\cite{walmer2023teaching} show that the
depth at which useful information concentrates in a ViT depends on
its supervision.
El~Banani~\etal~\cite{elbanani2024probing} probe single-view
surface reconstruction (depth, normals) and multiview correspondence,
finding that foundation models encode surface properties well but are
not 3D-consistent across large viewpoint changes. 
Tang~\etal~\cite{tang2023emergent} and
Zhang~\etal~\cite{zhang2023tale} show that diffusion features alone,
and DINO combined with diffusion features, both substantially improve
semantic correspondence over DINO alone.
Most
recently, You~\etal~\cite{you2025multiview} evaluate the 3D awareness
of several of the same backbones we study, but through a full
downstream one-shot pose-estimation pipeline: correspondences are
established via feature matching and then passed to a RANSAC-PnP
solver, and only the resulting pose accuracy is reported. 

\noindent\textbf{Novel object pose estimation.}
A growing applied literature estimates the 6D pose of objects unseen
during training by matching image features against 
rendered templates, motivating the retrieval setting we
study; see Liu~\etal~\cite{liu2026survey} for a comprehensive survey
of this broader area. GigaPose~\cite{nguyen2024gigapose} and its
predecessor~\cite{nguyen2022templates} establish template-based
retrieval as a fast, generalizable alternative to render-and-compare
pipelines such as MegaPose~\cite{labbe2022megapose}. 
SAM-6D~\cite{lin2024sam6d} and
OnePose~\cite{sun2022onepose} explore complementary
segmentation-driven and structure-from-motion-driven variants of the
same unseen-object setting. Goodwin~\etal~\cite{goodwin2022zeroshot}
address the related zero-shot category-level pose problem. Most
notably, FoundPose~\cite{ornek2024foundpose} builds its coarse pose
estimator directly on DINOv2 patch descriptors 
and observes that an
intermediate DINOv2 layer gives the most reliable
correspondences among a small set of evaluated layers. 
Most methods treat the choice of backbone, layer,
and model scale as a convention or justify it with a brief ablation.
   \section{Method}
\label{sec:method}

\begin{figure*}[t]
  \centering
  \includegraphics[width=0.9\linewidth]{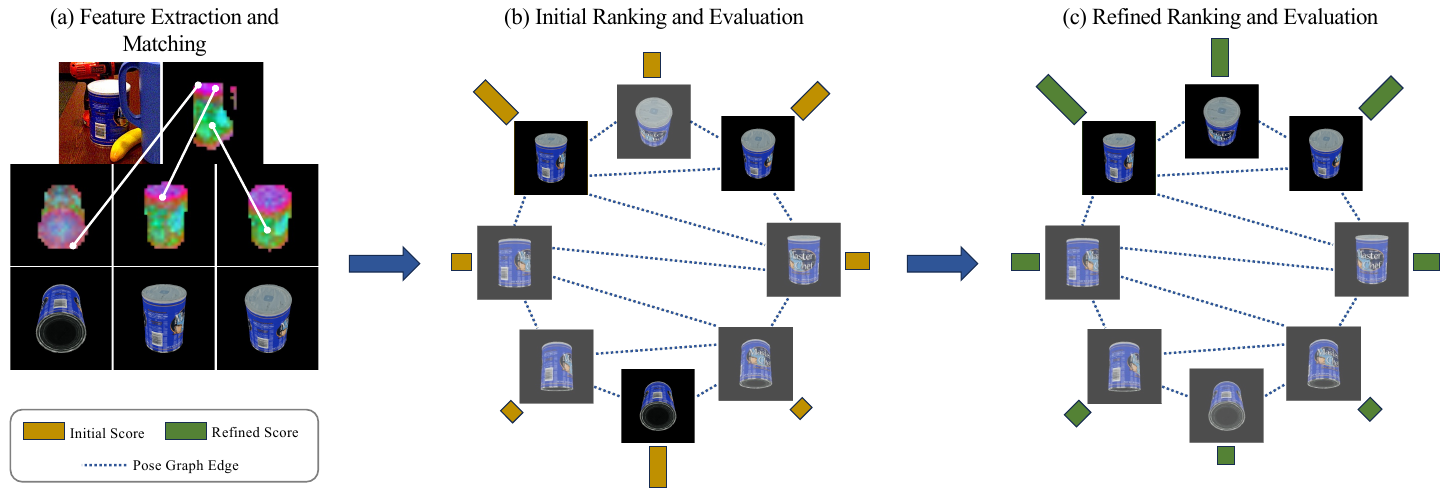}
  \caption{Benchmark pipeline. A query and a known-pose template
  gallery are independently mapped to masked local features
  (\cref{sec:feat}); their similarity yields a raw retrieval ranking
  $S^{(0)}$ (\cref{sec:retrieval}), which is optionally refined into
  $S^{(1)}$ using only the template-to-template pose graph
  (\cref{sec:refine}) before both are scored against the graded
  ground-truth similarity $S^\ast$ (\cref{sec:metrics-method}).}
  \label{fig:pipeline}
\end{figure*}

We
formalize \emph{pose-awareness} as follows: a representation is
pose-aware to the extent that feature similarity intrinsically
reflects the relative pose of an object, such that observations with
similar poses are nearby in feature space and observations with
increasingly different poses become correspondingly less similar. 
As is standard in
modern pose-estimation pipelines~\cite{liu2026survey}, the target object is assumed to be
detected and cropped to an approximately centered observation before
any feature is extracted, which substantially reduces the influence
of image-plane translation on feature similarity, so we treat
rotation as the primary remaining source of pose variation. 
Unlike a full
downstream pose-estimation pipeline~\cite{you2025multiview}, which couples representation
quality to a correspondence-sampling strategy and a geometric solver,
our framework
isolates the representation itself from any solver built on top of
it. \Cref{fig:pipeline} summarizes the resulting benchmark pipeline.

\subsection{Feature extraction}
\label{sec:feat}
For an RGB image $x$ with segmentation mask $M$, a frozen backbone
$\Phi$ produces a spatial grid of patch tokens; any prefix tokens
(\eg~\texttt{[CLS]} or register tokens) are discarded, and the
remaining $N_p$ tokens are reshaped to their native $H\times W$
spatial grid. We probe each backbone at four \emph{relative} depths
$\ell \in \{0.20, 0.45, 0.70, 1.00\}$ of its total depth $D$ (block
index $\mathrm{round}(\ell D)$).
$M$ is resized to the
feature-grid resolution 
and used to
retain only foreground token locations, each subsequently
$\ell_2$-normalized:
\begin{equation}
  F(x) = \left\{\, \hat\phi_i = \phi_i / \lVert \phi_i \rVert_2 \;:\; i \in \mathrm{fg}(M) \,\right\},
  \label{eq:featset}
\end{equation}
where $\phi_i$ is the patch token at spatial location $i$ and
$\mathrm{fg}(M)$ denotes foreground locations. $F(x)$ is therefore an
unordered set of $\lvert \mathrm{fg}(M)\rvert$ unit-norm local
descriptors.

\subsection{Template retrieval}
\label{sec:retrieval}
Each object is represented by a gallery of $N_T{=}162$ rendered
templates $\{t_1,\dots,t_{N_T}\}$ at known ground-truth rotations
$\{R_1,\dots,R_{N_T}\} \subset SO(3)$.

\noindent\textbf{Feature similarity.}
For a query $q$ and template $t_j$, we compute a directional,
thresholded local nearest-neighbor similarity over their local
descriptor sets $F(q)$ and $F(t_j)$:
\begin{align}
  m(f,t_j) &= \max_{g \in F(t_j)} f^\top g, \label{eq:localmax}\\
  F_\tau(q,t_j) &= \{ f \in F(q) : m(f,t_j) \geq \tau \}, \label{eq:threshset}\\
  s(q,t_j) &= \operatorname*{mean}_{f \in F_\tau(q,t_j)} m(f,t_j), \label{eq:featsim}
\end{align}
i.e.\ every query patch descriptor is matched to its closest
template patch descriptor, matches below
$\tau{=}0.5$ are discarded, and the remaining scores are averaged.
Applying \cref{eq:featsim} for every template yields a feature
similarity vector $s(q,\cdot) \in \mathbb{R}^{N_T}$; stacking over a
query set of size $Q$ yields the similarity matrix $S \in
\mathbb{R}^{Q \times N_T}$ used for retrieval.

\noindent\textbf{Ground-truth pose similarity.}
We convert the $SO(3)$ geodesic distance between two rotations,
\begin{equation}
  d(R_a, R_b) = \arccos\!\left(\frac{\operatorname{tr}(R_a^\top R_b) - 1}{2}\right) \in [0,\pi],
  \label{eq:geodesic}
\end{equation}
into a bounded ground-truth similarity
\begin{equation}
  s^\ast(R_a,R_b) = \mathrm{clip}\!\left(1 - d(R_a,R_b)/\pi,\, 0,\, 1\right) \in [0,1].
  \label{eq:truesim}
\end{equation}
For objects
with rotational symmetry, we replace $d$ with the minimum distance
over symmetry-equivalent poses: for a discrete symmetry
group $\{S_n\}$ (identity included),
\begin{equation}
  d_{\text{sym}}(R_a,R_b) = \min_n d(R_a,\, R_b S_n);
  \label{eq:dsym}
\end{equation}
for continuous axial symmetry with axis
$\mathbf{u}$, rotation about $\mathbf{u}$ is ignored and the minimum angle between
symmetry-transformed axes is used (combined with any discrete
symmetries present). This yields a ground-truth similarity matrix
$S^\ast \in [0,1]^{Q\times N_T}$ that serves as the graded
relevance signal for evaluation (\cref{sec:metrics-method}) and, when
computed between templates, as the pose graph for
refinement (\cref{sec:refine}).

\noindent\textbf{Retrieval.}
For a query $q$, templates are ranked by descending $s(q,\cdot)$, and
the Top-$K$ retrieval is the index set of the $K$ highest-scoring
templates, for $K \in \{1,3,5,10\}$.

\subsection{Graph-based refinement}
\label{sec:refine}
Raw feature similarity (\cref{eq:featsim}) is computed independently
per query-template pair without utilizing the known relative poses of templates.
We therefore
propagate scores across a template-to-template pose graph before
final ranking. Let $A \in [0,1]^{N_T\times N_T}$, $A_{ij} =
s^\ast(R_i, R_j)$, be the ground-truth (symmetry-aware) similarity
between templates $i$ and $j$ computed with 
\cref{eq:geodesic}, but now over the fully known template
poses, requiring no additional supervision. We sparsify $A$ by
retaining, for each template $i$, only its $k{=}8$ most-similar
neighbors (each template's self-loop is always retained, $A_{ii}{=}1$)
and row-normalize into a transition matrix
\begin{equation}
  P = A \,/\, (A\mathbf{1}),
  \label{eq:transition}
\end{equation}
where $A\mathbf{1}$ denotes each row sum. Given the raw
feature-similarity matrix $S^{(0)}{=}S$ from \cref{eq:featsim}, one
step of personalized diffusion~\cite{zhou2004learning,iscen2017efficient} produces the refined score matrix
\begin{equation}
  S^{(1)} = (1-\alpha)\, S^{(0)} + \alpha\, S^{(0)} P^\top,
  \label{eq:refine}
\end{equation}
with smoothing strength $\alpha{=}0.2$.
We report all metrics
both before ($S^{(0)}$) and after ($S^{(1)}$) this step. Because $A$
is built from known ground truth
, this step is informative beyond raw
accuracy: if it improves an imperfect
$S^{(0)}$, the representation's errors are 
compatible with the underlying pose manifold; if not, those
errors are closer to unstructured noise and hard to correct through geometric
reasoning alone. 

\subsection{Evaluation metrics}
\label{sec:metrics-method}
Retrieval quality is scored against the graded ground-truth relevance
$S^\ast(q,\cdot)$ (\cref{sec:retrieval}). Let $\pi_q$ denote the
ranking of templates induced by the feature similarity $s(q,\cdot)$
(before or after refinement), and let $r_q(t) \in
\{1,\dots,N_T\}$ denote the resulting rank position of
template $t$ (so $r_q(t){=}1$ for the top-ranked template, among
all $N_T$ templates, not just those inside a Top-$K$ cutoff). We
write $\mathrm{Pred}_K(q)$ for the predicted Top-$K$ set under
$\pi_q$, and $\mathrm{True}_K(q)$ for the ideal Top-$K$ set
obtained by sorting templates by $S^\ast(q,\cdot)$.
The graded relevance of the template ranked $i$-th under $\pi_q$ is
\begin{equation}
  \mathrm{rel}_i = S^\ast(q, \pi_q(i)),
  \label{eq:rel}
\end{equation}
which the metrics below use repeatedly. 


\noindent\textbf{Graded rank quality (nDCG).}
Using the relevance scores from \cref{eq:rel}, we compute the
standard normalized discounted cumulative gain
\cite{jarvelin2002cumulated}
\begin{equation}
  \mathrm{nDCG}@K(q) = \frac{\mathrm{DCG}@K(q)}{\mathrm{IDCG}@K(q)},
  \label{eq:ndcg}
\end{equation}
\begin{equation}
  \mathrm{DCG}@K(q) = \sum_{i=1}^K \frac{\mathrm{rel}_i}{\log_2(i+1)},
  \label{eq:dcg}
\end{equation}
where $\mathrm{IDCG}@K(q)$ is $\mathrm{DCG}@K(q)$ under the ideal
ranking (templates sorted by $S^\ast(q,\cdot)$ directly, \ie\ the
same ordering that defines $\mathrm{True}_K(q)$ above). nDCG rewards
placing highly-relevant templates near the top of the
retrieval, not merely present in it.
Because \cref{eq:ndcg} is well-defined for every object
regardless of symmetry type, we use nDCG@$K$ as our primary metric
throughout \cref{sec:experiments}.

\noindent\textbf{Catastrophic failure rate.}
\begin{equation}
  \mathrm{Bad}@K(q) = \frac{1}{K} \sum_{i=1}^K \mathbf{1}\!\left[\mathrm{rel}_i < \tau_{\text{bad}}\right],
  \quad \tau_{\text{bad}} = 0.5,
  \label{eq:bad}
\end{equation}
\ie\ the fraction of the top-$K$ retrieval whose ground-truth
similarity falls below $0.5$ (more than $90^\circ$ from the true
pose). Unlike nDCG, which is a graded, continuous score, Bad@$K$
answers a binary question per retrieved template: is this an
match with a large pose error, or merely an imprecise one? It is
therefore more sensitive to the rate of severe retrieval failures
than to small ranking imperfections. This threshold-based indicator
is adapted to our pose-similarity setting and is not itself drawn
from prior work.

\noindent\textbf{Additional metrics.}
Beyond nDCG and Bad@$K$, we report seven further metrics: Hit@$K$, MRR, Top-$K$ overlap, ValidPrec@$K$, ValidRecall@$K$,
Mean rank, and Median rank, built
directly from $\mathrm{Pred}_K(q)$, $\mathrm{True}_K(q)$ (or, for
symmetric objects, a tolerance-based valid-match set $V(q)$, and
$r_q$). We formally define the remaining seven metrics in the supplementary
material; the main paper reports nDCG and Bad@$K$ throughout, \ie\
the two metrics that are well-defined for every object regardless of symmetry type and that correlate strongly with the remaining seven
(see the supplementary material for the full cross-metric analysis).
All per-query metrics are averaged over queries, then macro-averaged over datasets.

   \section{Experiments}
\label{sec:experiments}


\subsection{Implementation details}
\label{sec:impl}

\noindent\textbf{Models.}
We evaluate 31 publicly available checkpoints spanning 10 pretraining
families and five forms of supervision: self-distillation
(DINO~\cite{caron2021emerging}, DINOv2~\cite{oquab2023dinov2}, 
DINOv2 with register tokens~\cite{darcet2024vitneedreg}, DINOv3~\cite{simeoni2025dinov3}), masked-image reconstruction
(MAE~\cite{he2022mae}, BEiT~\cite{bao2022beit}), CLIP-style
vision--language contrastive learning~\cite{radford2021clip},
masked reconstruction distilled from a CLIP teacher
(EVA-02~\cite{fang2023eva02}), and supervised classification
(the original ViT recipe~\cite{dosovitskiy2021vit} and
DeiT~III~\cite{touvron2022deit3}). Checkpoints range from 21M
(DINOv3-Small) to 1.1B parameters (DINOv2-Giant); we group them into five
comparable size tiers (Small/Base/Large/Huge+/Giant) for
cross-family comparison. All backbones are loaded through
\texttt{timm}~\cite{wightman2019pytorch}. For every backbone we extract patch tokens at four
\emph{relative} network depths: approximately 20\%, 45\%, 70\%, and
100\% (i.e.\ the final block), so that models with different total
depths (12, 24, 32, or 40 transformer blocks) remain comparable on a
common depth axis.

\noindent\textbf{Retrieval and refinement pipeline.}
Two query regimes are evaluated: \textbf{templates},
where the query is itself a held-out rendered template (excluding the
self-match), and \textbf{real images}, where the query is a real
photograph of the object. The templates regime holds appearance
nuisance factors (background, illumination, sensor noise, occlusion)
essentially fixed, isolating pure pose understanding. 
The real-images regime instead introduces the
full sim-to-real domain gap.
More details are available in Supplementary. 

\noindent\textbf{Datasets.}
We construct PoseBench based on seven core BOP Benchmark~\cite{hodan2018bop, sundermeyer2023bop} datasets, including
HB, IC-BIN, ITODD, LM-O, T-LESS, TUD-L, and YCB-V, which cover 125
objects in total (\cref{tab:datasets}). Objects are 
annotated with a symmetry type (asymmetric, discrete, continuous, or
discrete+continuous), used by the ground-truth pose-similarity
computation to treat symmetry-equivalent templates as equally
correct.
We use real images from the test sets, except for HB and ITODD, where ground-truth test poses are unavailable and validation images are used instead.
More details are provided in Supplementary.

\begin{table}[t]
  \centering
  \small
  \setlength{\tabcolsep}{3.2pt}
  \begin{tabular}{@{}lccccccc@{}}
    \toprule
    & HB & IC-B & ITD & LM-O & T-L & TUD-L & YCB-V \\
    \midrule
    \#Obj. & 33 & 2 & 28 & 8 & 30 & 3 & 21 \\
    \% sym. & 9 & 50 & 64 & 25 & 90 & 0 & 33 \\
    \#Query & 4621 & 2186 & 123 & 1514 & 6873 & 600 & 4125 \\
    \bottomrule
  \end{tabular}
  \caption{Number of objects, the fraction with rotational symmetry
  (discrete or continuous), and the number of real-image query
  instances, per BOP dataset (125 objects, 20{,}042 real-image
  queries, and 20{,}250 rendered templates with exactly 162 per object
  in every dataset). Object diameters range from 24mm
  (smallest ITODD part) to 478mm (largest HB object) across the
  benchmark.}
  \label{tab:datasets}
\end{table}

\noindent\textbf{Evaluation metrics.}
Because nDCG@$K$
(\cref{eq:ndcg}) is the only metric with an identical definition for
all object types, we use it as our primary metric throughout; we
verified that it agrees well with Bad@$K$ (\cref{eq:bad};
$\rho{=}0.97$--$0.99$) but only moderately with Hit@$K$ in the
templates regime ($\rho{=}0.50$), since Hit@$K$ saturates once most models place \emph{some} correct template in the top-$K$ (see
in Supplementary for the full cross-metric analysis). 
Unless noted, we report results \emph{macro-averaged}: the mean is first taken over objects within each of the 7 datasets, and then over the 7 resulting dataset-level means. This avoids the two large datasets (HB, T-LESS)
dominating small datasets.


\begin{figure}[t]
  \centering
  \includegraphics[width=0.85\linewidth]{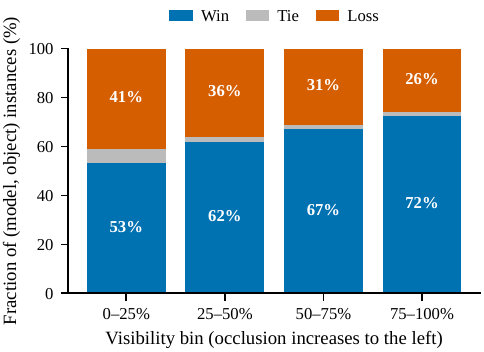}
  \caption{Refinement reliability (win/tie/loss rate, real images,
nDCG@5) drops sharply with occlusion: it helps 72\% when
the object is mostly visible, but only 53\% when it is mostly
occluded.}
  \label{fig:refinement}
\end{figure}

\subsection{Quantitative results}
\label{sec:quant}

\noindent\textbf{Graph-based refinement.}
Refinement improves all nine metrics, in both query regimes, with
every improvement statistically significant (paired Wilcoxon
signed-rank test over the 125 objects, $p{<}0.001$;
\cref{tab:refinement}).

\begin{figure}[t]
  \centering
  \includegraphics[width=\linewidth]{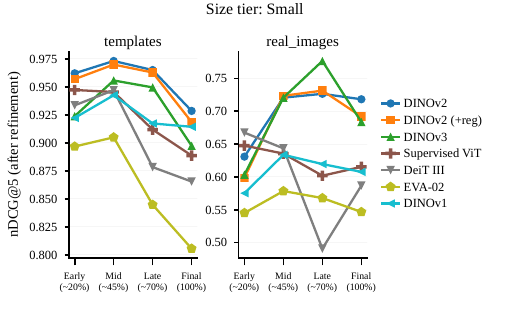}
  \caption{nDCG@5 (after refinement) vs.\ relative network depth,
  Small-size models only.}
  \label{fig:layerdepth-small}
\end{figure}

\begin{figure}[t]
  \centering
  \includegraphics[width=\linewidth]{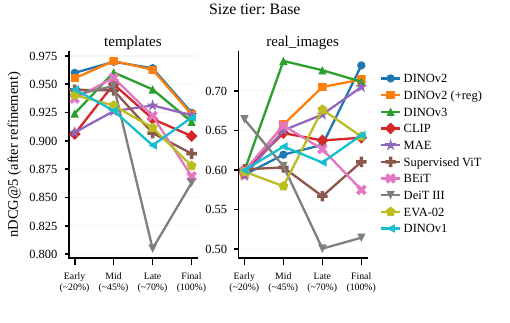}
  \caption{nDCG@5 (after refinement) vs.\ relative network depth,
  Base-size models only.}
  \label{fig:layerdepth-base}
\end{figure}

\begin{figure}[t]
  \centering
  \includegraphics[width=\linewidth]{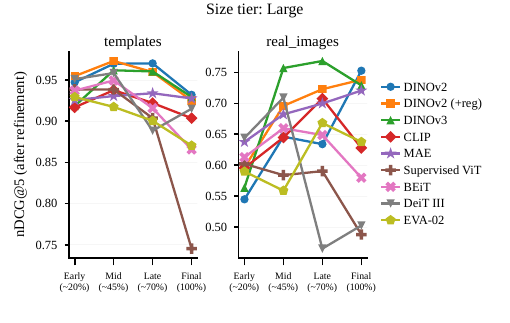}
  \caption{nDCG@5 (after refinement) vs. relative network depth for Large-size models. 
  }
  \label{fig:layerdepth-large}
\end{figure}


\begin{table*}[t]
  \centering
  \small
  \setlength{\tabcolsep}{5pt}
  \begin{tabular}{@{}llcccccc@{}}
    \toprule
    Query type & Metric & N obj. & Before & After & $\Delta$ (abs.) & \% improvement & Sig. \\
    \midrule
    Templates & nDCG@5 & 125 & 0.9124 & 0.9273 & +0.0148 & +1.6\% & $p{<}0.001$ \\
    Templates & Hit@5 & 125 & 0.8872 & 0.9077 & +0.0204 & +2.3\% & $p{<}0.001$ \\
    Templates & MRR & 125 & 0.5331 & 0.5554 & +0.0223 & +4.2\% & $p{<}0.001$ \\
    Templates & Top-$K$ overlap & 67 & 0.5953 & 0.6309 & +0.0356 & +6.0\% & $p{<}0.001$ \\
    Templates & ValidPrec@5 & 58 & 0.3873 & 0.4071 & +0.0198 & +5.1\% & $p{<}0.001$ \\
    Templates & ValidRecall@5 & 58 & 0.3987 & 0.4426 & +0.0439 & +11.0\% & $p{<}0.001$ \\
    Templates & Bad@5 & 125 & 0.0773 & 0.0634 & $-$0.0139 & +18.0\% & $p{<}0.001$ \\
    Templates & Mean rank & 125 & 9.94 & 8.70 & $-$1.24 & +12.5\% & $p{<}0.001$ \\
    Templates & Median rank & 125 & 8.31 & 7.34 & $-$0.96 & +11.6\% & $p{<}0.001$ \\
    \midrule
    Real images & nDCG@5 & 125 & 0.6305 & 0.6405 & +0.0100 & +1.6\% & $p{<}0.001$ \\
    Real images & Hit@5 & 125 & 0.4000 & 0.4074 & +0.0074 & +1.9\% & $p{<}0.001$ \\
    Real images & MRR & 125 & 0.1997 & 0.2074 & +0.0077 & +3.8\% & $p{<}0.001$ \\
    Real images & Top-$K$ overlap & 67 & 0.1617 & 0.1723 & +0.0107 & +6.6\% & $p{<}0.001$ \\
    Real images & ValidPrec@5 & 58 & 0.1674 & 0.1750 & +0.0076 & +4.5\% & $p{<}0.001$ \\
    Real images & ValidRecall@5 & 58 & 0.1616 & 0.1707 & +0.0091 & +5.6\% & $p{<}0.001$ \\
    Real images & Bad@5 & 125 & 0.4416 & 0.4282 & $-$0.0134 & +3.0\% & $p{<}0.001$ \\
    Real images & Mean rank & 125 & 51.29 & 48.61 & $-$2.68 & +5.2\% & $p{<}0.001$ \\
    Real images & Median rank & 125 & 49.60 & 47.01 & $-$2.58 & +5.2\% & $p{<}0.001$ \\
    \bottomrule
  \end{tabular}
  \caption{Effect of graph-based refinement on all 9 evaluated metrics
  ($K{=}5$, macro-averaged). \% improvement is the relative change
  toward the better direction for each metric (for Bad@5, Mean rank,
  and Median rank, where lower is better, a positive value reflects
  a reduction). Significance is a paired Wilcoxon signed-rank test~\cite{wilcoxon1945individual} over
  the 125 objects.}
  \label{tab:refinement}
\end{table*}

Putting every metric on this common relative scale (\cref{tab:refinement},
rightmost columns) shows that refinement's practical effect is
understated by the nDCG headline number alone: nDCG@5 improves by
only 1.4--1.6\%, but Bad@5 (the catastrophic-failure rate) drops by
17.8\% for templates, and mean/median rank improve by 11--12\%. This demonstrates that
refinement fixes a substantially larger share of 
retrievals with large pose errors than the nDCG number alone suggests, since nDCG is a
graded metric dominated by an already-high baseline.

Examining individual (model, object) instances rather than averages reveals an important asymmetry that aggregate results obscure: refinement improves nDCG@5 in 97.3\% of template cases and degrades it in only 1.0\% (restricted to asymmetric objects, the loss rate falls further, to 0.3\%). For real images, refinement improves nDCG@5 in only 73.2\% of cases, ties in 1.4\%, and actively degrades retrieval quality in 25.4\%.   This loss rate rises to 41\% under heavy occlusion (0 to 25\% visibility; \cref{fig:refinement}, left)).  This is because refinement
propagates through a graph seeded by the raw similarity scores; under
heavy occlusion or continuous symmetry, that seed is already too
unreliable for propagation to recover. 
This pattern reveals a specific property about the representations: their
pose-relevant errors are predominantly local and can be generally mitigated when the correct pose neighbors are known, and shift toward unstructured noise mainly in the specific
regime (heavy occlusion) where the raw similarity signal itself has
already broken down.

\begin{figure}[t]
  \centering
  \includegraphics[width=\linewidth]{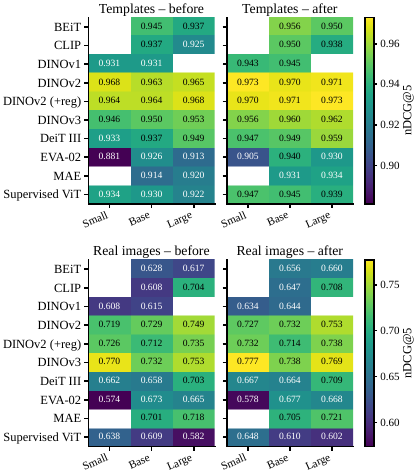}
  \caption{Family$\times$size (Small/Base/Large) nDCG@5, before/after
  refinement, templates vs.\ real images (own best layer).}
  \label{fig:heatmap}
\end{figure}

\noindent\textbf{Best model, layer, and size.}
\cref{fig:layerdepth-small,fig:layerdepth-base,fig:layerdepth-large}
show where pose-aware information is concentrated in the network,
separately for each size tier.
Across all three tiers, the templates regime consistently peaks at
$\sim$45\% relative depth (22 of 31 models), and the final
layer is essentially never best. The real-image regime instead
consistently peaks later, at $\sim$70\% depth or the final block (18
of 31 models). We interpret this as a trade-off between pose-specificity (sharpest mid-network) and domain-robustness (deeper
layers), and note that the location of this trade-off remains stable across
scale.
Restricting the family$\times$size grid to the three sizes present
across nearly every family (\cref{fig:heatmap}) shows DINOv2 with or without register tokens dominate the templates regime at every
size, while \mbox{DINOv3-Small} is the single best cell of the entire
real-image grid (nDCG@5 $=0.777$), ahead of DINOv3-Large. Model-size
scaling is \emph{not} monotonic: DINOv2-Giant (1.1B params, nDCG@5
$=0.771$ on real images) is dominated by DINOv3-Small (21M params,
nDCG@5 $=0.777$), a 50$\times$ reduction in parameters for a higher
score. Bootstrap confidence intervals ~\cite{tibshirani1993introduction} over the 125 objects confirm
this is not noise: the top handful of DINOv2/DINOv3 models at every
tested scale have heavily overlapping 95\% CIs and are not pairwise
distinguishable (paired Wilcoxon, $p{>}0.05$), while models several
ranks lower are clearly separated.
We provide a full numerical breakdown behind
\cref{fig:heatmap} in Supplementary.




\begin{table*}[t]
  \centering
  \small
  \setlength{\tabcolsep}{6pt}
  \begin{tabular}{@{}lccccc@{}}
    \toprule
    Family & nDCG templates & nDCG real images & Domain gap (abs., 95\% CI) & Domain gap (rel.) \\
    \midrule
    DINOv2 & 0.932 & 0.736 & 0.195~[0.170, 0.221] & 21.0\% \\
    DINOv2 (+reg) & 0.926 & 0.726 & 0.200~[0.175, 0.225] & 21.6\% \\
    MAE & 0.925 & 0.713 & 0.213~[0.191, 0.235] & 23.0\% \\
    DINOv3 & 0.912 & 0.692 & 0.220~[0.199, 0.242] & 24.1\% \\
    EVA-02 & 0.851 & 0.609 & 0.243~[0.220, 0.266] & 28.5\% \\
    CLIP & 0.904 & 0.634 & 0.270~[0.243, 0.297] & 29.8\% \\
    Supervised ViT & 0.841 & 0.571 & 0.270~[0.240, 0.300] & 32.1\% \\
    BEiT & 0.867 & 0.577 & 0.290~[0.266, 0.315] & 33.4\% \\
    DINOv1 & 0.918 & 0.626 & 0.292~[0.266, 0.318] & 31.8\% \\
    DeiT~III~$\dagger$ & 0.879 & 0.524 & 0.355~[0.328, 0.383] & 40.4\% \\
    \bottomrule
  \end{tabular}
  \caption{Template-only vs.\ real-image nDCG@5 by family, at the last layer, after refinement, sorted by increasing domain gap. The 95\% CI on the absolute gap comes from clustered bootstrap resampling (5{,}000 iterations), resampling objects within each of the 7 datasets before averaging, following the macro-averaging protocol of Sec.~4.3. $\dagger$ One of the four DeiT~III checkpoints (Large) is evaluated at 384px rather than 224px. Excluding it still leaves a gap of 0.336, the largest of all 10 families.}
  \label{tab:domaingap}
\end{table*}

\noindent\textbf{Domain gap.}
To evaluate how performance degrades away from the
clean, controlled templates regime, we start with the sim-to-real domain gap.
\Cref{tab:domaingap} reports, per family, nDCG@5 on templates against
real images and the resulting domain gap, both at the last
layer.
DINOv2 has the smallest
average relative domain-gap (21.0\%, 95\% CI [17.0,22.1]) and
DeiT~III the largest (40.4\%, significantly worse than every other
family); this ranking is robust to the fact that one DeiT~III
checkpoint is evaluated at 384px rather than 224px (excluding it
still leaves DeiT~III worst). The gap is also strongly modulated by
object symmetry, rising from an 11.4\% relative loss for the most
symmetric objects to 31.0\% for fully asymmetric ones, since symmetric
objects retain a broader equivalent template matches. A cross-metric check shows DeiT~III's last-place
ranking is metric-invariant, but DINOv2's first-place ranking is not:
by mean/median rank, MAE has a smaller gap than DINOv2, so we report
DINOv2 as the most \emph{nDCG-robust} family.

\begin{figure}[t]
  \centering
  \includegraphics[width=\linewidth]{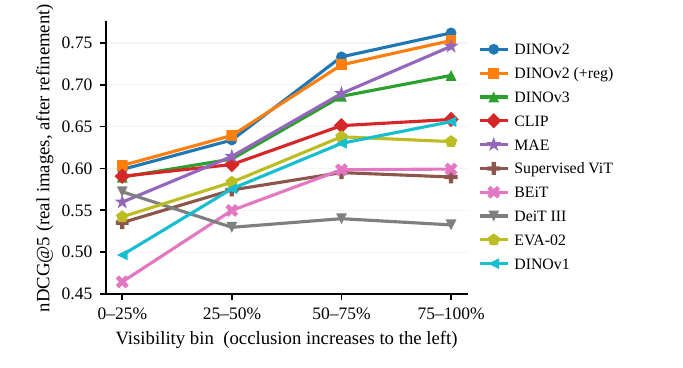}
  \caption{nDCG@5 (real images, after refinement) vs.\ visibility bin,
  by family.}
  \label{fig:occlusion}
\end{figure}

\noindent\textbf{Robustness to occlusion.}
\label{sec:occlusion}
We examine the degradation due to occlusion within
the real-images regime by binning queries by visible-surface fraction
(\cref{fig:occlusion}): DINOv2 with or without register tokens degrade
the least from full to $[0,0.25)$ visibility
, while BEiT and Supervised~ViT degrade the most under heavy
occlusion in absolute terms. DeiT~III shows almost no degradation, but
this is a \emph{floor effect}, not genuine robustness, since its
absolute performance (0.53--0.57) is uniformly low across all
visibility levels.

\subsection{Qualitative analysis}
\label{sec:qual}
Beyond the aggregate trends above, three object and dataset level
patterns are worth highlighting qualitatively. First, difficulty is
concentrated rather than uniform: ITODD (industrial, largely
textureless, thin-structured parts, and captured as grayscale rather than color) is the
hardest dataset in the templates regime (nDCG@5 $=0.881$) for
\emph{every} model we tested, indicating an intrinsic property of the
objects and sensor modality rather than a model-specific weakness.
Second,
an oracle that selects the best-performing model \emph{per object}
rather than committing to one fixed model closes a gap of $+0.020$
nDCG on templates but $+0.066$ on real images, which is three times more
unresolved headroom in the realistic setting. Third, the model chosen most often by that oracle is not
one of the leaderboard-topping DINOv2/DINOv3 variants: MAE-ViT-L/16
wins 41 of 125 objects on templates and 30 of 125 on real images,
concentrated specifically in the HB and T-LESS datasets, despite
ranking only $\sim$13th on average nDCG. We verified this is not a
near-ceiling tie artifact, since many of MAE's per-object wins occur on
objects with nDCG in the 0.78--0.90 range, well below saturation. This
suggests masked-reconstruction features capture appearance cues that
self-distillation objectives under-weight for specific object
categories, a concrete motivation for future feature-fusion work. 
More qualitative analysis is available in Supplementary.

   \section{Conclusion}
\label{sec:conclusion}
We introduced PoseBench, a systematic benchmark that isolates the
pose-awareness of visual foundation model representations from any
downstream pose-estimation pipeline, and used it to evaluate 31
checkpoints spanning 10 pretraining families and 5 size tiers. Across
this benchmark, pose awareness presents broadly after large-scale
visual pretraining even without explicit 3D supervision. 
Its presence is
non-uniform,
and consistent gains from pose-graph refinement across every occlusion regime indicate
that even imperfect feature similarities remain compatible with the
underlying pose manifold.
Together, these results replace convention-driven
assumptions in the applied pose-estimation literature,
with an evidence-based, statistically validated account of where, and
how reliably, pose awareness lives in frozen visual representations.

{
    \small
    \bibliographystyle{ieeenat_fullname}
    \bibliography{main}
}

\clearpage
\maketitlesupplementary

This supplementary material accompanies the main paper. It is
organized into five sections: \cref{sec:supp-metrics} formally defines the
seven additional evaluation metrics named but not defined in the
main paper's Sec.~3.4; \cref{sec:supp-impl-details} provides
additional implementation details for the retrieval and refinement
pipeline, including the datasets used and the domain-gap evaluation
methodology; \cref{sec:supp-heatmaps} provides the full
family $\times$ size numerical breakdown behind the main paper's
Fig.~8; \cref{AQA} gives an additional qualitative
analysis of dataset-level effects not captured by the aggregate
metrics; and \cref{sec:supp-discussion} gives an extended discussion,
limitations, and future work.

\appendix
\section{Additional Evaluation Metrics}
\label{sec:supp-metrics}

The main paper (Sec.~3.4) defines our two primary metrics, nDCG@$K$
and Bad@$K$, which are graded and well-defined for every object
regardless of symmetry type. Here we formally define the seven
remaining metrics named in the main paper: Top-$K$ overlap, Hit@$K$,
MRR, ValidPrec@$K$, ValidRecall@$K$, Mean rank, and Median rank.
Because no single template is uniquely correct for a symmetric
object, four of these metrics, Hit@$K$, MRR, Mean rank, and Median
rank, need a different formula depending on object type: we give
both below, and report whichever applies to each object. Top-$K$
overlap, ValidPrec@$K$, and ValidRecall@$K$ are each defined for one
object type only.

\paragraph{Asymmetric objects: rank-based metrics.}
For the 67 of 125 objects with no rotational symmetry, a single
best-matching template is well-defined, so we report four further
metrics built directly from $\mathrm{Pred}_K(q)$,
$\mathrm{True}_K(q)$, and $r_q$:
\begin{align}
  \text{Top-}K\text{ overlap}(q) &= \frac{|\mathrm{Pred}_K(q) \cap \mathrm{True}_K(q)|}{K}, \label{eq:overlap}\\
  \mathrm{Hit}@K(q) &= \mathbf{1}\!\left[\mathrm{Pred}_K(q) \cap \mathrm{True}_K(q) \neq \emptyset\right], \label{eq:hit}\\
  \mathrm{MRR}(q) &= \frac{1}{r_q(t^\ast)}, \quad t^\ast = \operatorname*{arg\,max}_t S^\ast(q,t), \label{eq:mrr}\\
  \overline{r}(q), \widetilde{r}(q) &= \operatorname*{mean}_{t \in \mathrm{True}_K(q)} r_q(t),\; \operatorname*{median}_{t \in \mathrm{True}_K(q)} r_q(t). \label{eq:meanrank}
\end{align}
Top-$K$ overlap (\cref{eq:overlap}) measures \emph{how much} of the
ideal top-$K$ was recovered; Hit@$K$ (\cref{eq:hit}), equivalently
the hit ratio used for top-$K$ retrieval evaluation by
He~\etal~\cite{he2017ncf}, relaxes this to a binary ``recovered at
least one'' signal. MRR (\cref{eq:mrr}), the mean reciprocal rank of
Voorhees~\cite{voorhees1999trec8}, does not depend on $K$ at all --
it is the reciprocal rank, among \emph{all} $N_T$ templates, of the
single best-matching template $t^\ast$, and is dominated by whether
that one template is found near the top of the full ranking. Mean
and median rank (\cref{eq:meanrank}), following the mean-rank
ranking-evaluation protocol of Bordes~\etal~\cite{bordes2013transe},
by contrast describe where the \emph{whole} ideal top-$K$ set ends
up in the predicted ranking on average.

\paragraph{Symmetric objects: valid-match analogues.}
For the remaining 58 objects (discrete, continuous, or
discrete+continuous symmetry), no single template is uniquely
correct, so we replace $\mathrm{True}_K(q)$ with the \emph{valid-match
set}
\begin{equation}
  V(q) = \left\{\, t : S^\ast(q,t) \geq \tau_{\text{valid}} \,\right\},
  \quad \tau_{\text{valid}} = 1 - \frac{15^\circ \cdot \pi / 180^\circ}{\pi},
  \label{eq:validset}
\end{equation}
\ie\ every template within a $15^\circ$ geodesic tolerance of the
query's true pose after accounting for symmetry, generally
containing more than one template and of variable size across
queries. Substituting $V(q)$ for $\mathrm{True}_K(q)$ (and, for the
rank-based pair, taking the \emph{best} rank over the whole set
rather than a single $t^\ast$) gives Hit@$K$'s, MRR's, and mean/median
rank's symmetric-object formulas:
\begin{align}
  \mathrm{HitAny}@K(q) &= \mathbf{1}\!\left[\mathrm{Pred}_K(q) \cap V(q) \neq \emptyset\right], \\
  \mathrm{MRRAny}(q) &= \frac{1}{\min_{t \in V(q)} r_q(t)}, \\
  \overline{r}_V(q), \widetilde{r}_V(q) &= \operatorname*{mean}_{t \in V(q)} r_q(t),\; \operatorname*{median}_{t \in V(q)} r_q(t),
\end{align}
and the same substitution defines the symmetric-only metrics
\begin{align}
  \mathrm{ValidPrec}@K(q) &= \frac{|\mathrm{Pred}_K(q) \cap V(q)|}{K}, \\
  \mathrm{ValidRecall}@K(q) &= \frac{|\mathrm{Pred}_K(q) \cap V(q)|}{|V(q)|}.
\end{align}
ValidPrec@$K$ and ValidRecall@$K$ separate the same two questions as
Top-$K$ overlap does for asymmetric objects: precision asks what
fraction of the retrieval is correct, recall asks what fraction of
all valid poses was found. All per-query metrics above are averaged
over queries, then macro-averaged over datasets as described in the
main paper (Sec.~4.3).

\paragraph{Cross-metric agreement.}
We verified that nDCG@$K$ agrees well with Bad@$K$
($\rho{=}0.97$--$0.99$ across all model, layer, and query-regime
configurations) but only moderately with Hit@$K$ in the templates
regime ($\rho{=}0.50$), since Hit@$K$ saturates once most models
place some correct template in the top-$K$, consistent with our
choice of nDCG@$K$ and Bad@$K$ as the two metrics reported in the
main paper.
\section{Additional Implementation Details}
\label{sec:supp-impl-details}

\paragraph{Retrieval and refinement pipeline.}
The main paper formalizes the full pipeline; here we specify the
concrete settings used in all experiments below. For every query
image we (i) extract patch features, (ii) mask out background patches
using the object's ground-truth segmentation mask, within a square, instance-centered
crop covering the visible object at roughly $1.5\times$ padding
, (iii) compute patch-level
nearest-neighbor similarity between the query and every one of the
object's 162 rendered templates (match threshold $\tau{=}0.5$,
, and (iv) rank templates by the aggregated
similarity to obtain the Top-$K$ retrieval for $K \in \{1,3,5,10\}$.
We additionally evaluate the optional graph-based refinement of
($\alpha{=}0.2$, $k{=}8$ pose-neighbors, one
diffusion step), reporting results both before and after this step.
Comparing the two regimes therefore separates whether a
representation contains intrinsic pose-aware structure from how
robustly that structure survives real-world appearance variation. 

\paragraph{Datasets.}
Real-image query sets are built by reorganizing each dataset's
scene-wise BOP annotations 
into an object-wise layout: every annotated instance of a given object
is
cropped from its source image using a square region centered on the
visible bounding box, padded to
$1.5\times$ the box's longer side.
Each instance's ground-truth pose and visible fraction are retained in per-instance metadata; the latter
defines the occlusion bins. 
Template galleries reuse the pre-rendered multi-view sets from
GigaPose~\cite{nguyen2024gigapose}, processed with the same
square-crop procedure.

\paragraph{Domain Gap.}
We report per family, nDCG@5 on templates against
real images and the resulting domain gap, both at the last
layer.
deliberately, rather than at each model's own best layer. 
Because the best layer itself differs
between the two regimes (previous paragraph), scoring each regime at
its own best layer would confound two effects at once: genuine
domain-gap sensitivity, and the fact that layer choice is itself
regime-dependent. Fixing both regimes to the same layer removes that
confound; we use the last layer specifically because it is a
neutral, architecture-defined reference point that was not selected
based on either regime's own performance, avoiding the circularity of
picking, \eg, the templates-best layer and thereby biasing the
comparison in templates' favor. Readers interested in the
own-best-layer picture instead should refer back to
the heatmaps
, which reports both regimes' nDCG@5 at each
model's own best layer side by side for Small/Base/Large models, from
which the corresponding gap can be read off directly. 
\section{Full Family\texorpdfstring{$\times$}{x}Size Breakdown}
\label{sec:supp-heatmaps}

The main paper's Fig.~8 summarizes nDCG@5 across the
family$\times$size grid (Small/Base/Large, own best layer, $K{=}5$),
before and after refinement, for both query regimes. Here we provide
the full numerical breakdown behind that figure: 2 refinement stages
(before/after) $\times$ 4 values of $K$ ($1$, $3$, $5$, $10$), for
both query regimes -- 16 heatmap panels in total, restricted to
Small/Base/Large as in the main paper. The family ordering is
essentially unchanged across all 16 panels, consistent with the
size-split depth curves in the main paper (Figs.~5--7); these two
grids are included for exact numerical reference rather than as a
required reading path.

\begin{figure*}[t]
  \centering
  \includegraphics[width=0.92\textwidth]{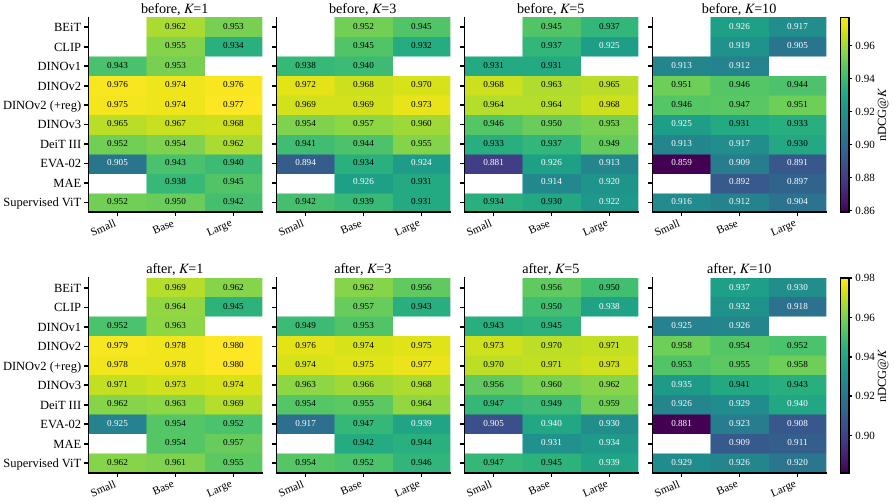}
  \caption{Family$\times$size (Small/Base/Large) nDCG@$K$, full stage
  $\times$ $K$ grid, \textbf{templates} query regime. Rows: before /
  after refinement. Columns: $K=1,3,5,10$.}
  \label{fig:s-full-grid-templates}
\end{figure*}

\begin{figure*}[t]
  \centering
  \includegraphics[width=0.92\textwidth]{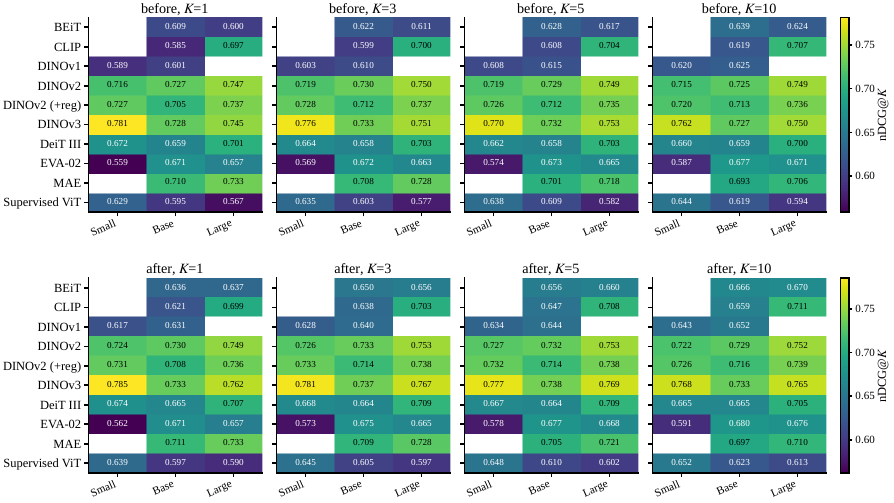}
  \caption{Family$\times$size (Small/Base/Large) nDCG@$K$, full stage
  $\times$ $K$ grid, \textbf{real-images} query regime. Rows: before /
  after refinement. Columns: $K=1,3,5,10$.}
  \label{fig:s-full-grid-real}
\end{figure*}

\section{{Additional Qualitative Analysis}}
\label{AQA}

T-LESS's apparently mild real-image
gap is partly a protocol artifact of its object
composition rather than genuinely easier retrieval, since 90\% of its objects (27/30) carry discrete or
continuous rotational symmetry, the highest fraction of any dataset
in the benchmark, versus 0\% for TUD-L.Because the symmetry-aware evaluation credits multiple templates as simultaneously ``correct'' for such objects, the score is mechanically inflated rather than reflecting genuinely easier retrieval. This pattern is not unique to T-LESS. Table~\ref{tab:dataset-symmetry} shows that every dataset's rank shift between the templates and real-image regimes tracks its symmetric-object fraction, with two exceptions among the datasets with very few objects. Itodd, the second-most symmetric dataset (64.3\% of objects), shows the same effect on a smaller scale: it is the single hardest dataset under the templates regime, yet only the fifth-hardest of seven under real images, the same direction of rank inversion seen in T-LESS. TUD-L, which is wholly asymmetric, shows the opposite pattern: easiest under templates but hardest under real images, consistent with the same mechanism running in reverse when no symmetry credit is available to cushion the domain gap.

\begin{table*}[ht]
\centering
\small
\begin{tabular}{lccccc}
\toprule
Dataset & \% Symmetric & nDCG@5 & nDCG@5 & Rank & Rank \\
        & objects      & (templ.) & (real) & (templ.) & (real) \\
\midrule
itodd          & 64.3\% & 0.856 & 0.657 & 1 (hardest) & 5 \\
lmo            & 25.0\% & 0.885 & 0.605 & 2           & 2 \\
ycbv           & 33.3\% & 0.898 & 0.671 & 3           & 6 \\
hb             &  9.1\% & 0.900 & 0.646 & 4           & 4 \\
tless          & 90.0\% & 0.900 & 0.735 & 5           & 7 (easiest) \\
icbin$^\dagger$& 50.0\% & 0.918 & 0.624 & 6           & 3 \\
tudl$^\dagger$ &  0.0\% & 0.921 & 0.588 & 7 (easiest) & 1 (hardest) \\
\bottomrule
\end{tabular}
\caption{Per-dataset symmetric-object fraction and difficulty rank under each query regime (rank 1 = hardest, own-best-layer, K=5, after refinement, DINOv3-qkvb checkpoints excluded). Rank inverts toward ``easier'' for high-symmetry datasets (itodd, T-LESS) and toward ``harder'' for the wholly-asymmetric TUD-L. $^\dagger$icbin (2 objects) and TUD-L (3 objects) have too few objects for the ranking to be statistically meaningful; both are included for completeness rather than as strong evidence.}
\label{tab:dataset-symmetry}
\end{table*}
\section{Discussion, Limitations, and Future Work}
\label{sec:supp-discussion}

\paragraph{Discussion.}
Our results extend the finding of El~Banani~\etal~\cite{elbanani2024probing}
that foundation-model representations, while encoding substantial
single-view surface information, are not multiview 3D-consistent: we
show this inconsistency manifests concretely as (i) a persistent
20--40\% relative performance drop whenever a real-world domain gap is
introduced, even for the strongest families, and (ii) a systematic
dependence on \emph{which} layer is used that itself changes with the
regime, contradicting the common default of extracting only the final
transformer block. Our refinement probe adds a third, more
constructive piece to that picture: the 3D-inconsistency
El~Banani~\etal\ identify is not, for the most part, unstructured
noise. Across nearly every model and condition we test, whatever
pose-relevant error a representation makes is locally coherent enough
that a graph built purely from known template geometry can correct
it, which is direct evidence that these representations retain
meaningful, if imperfect, pose structure rather than lacking it
outright. Practically, our results suggest that a
practitioner building a pose-aware retrieval system should prefer a
mid-size ($\sim$20--300M parameter) DINOv2 or DINOv3 checkpoint at
$\sim$45--70\% relative depth over a larger or final-layer alternative.
Graph-based score refinement is, on this evidence, worth applying
generally rather than reserving for near-ideal conditions: its
primary effect, an nDCG gain and a reduction in the catastrophic
Bad@$K$ failure rate, holds in every occlusion bin we tested,
including the most heavily occluded one, though not every metric
moves the same way in every bin (Hit@$K$'s aggregate value dips
slightly except under the heaviest occlusion), and its
\emph{instance-level} reliability still declines as occlusion
increases. A practitioner should expect a smaller and less consistent
benefit under heavy occlusion, not withhold the correction
altogether. Nor should one assume that any single model dominates
uniformly across object categories -- our oracle-ensemble analysis
suggests meaningful, currently untapped complementarity between
self-distillation and reconstruction-based features.

\paragraph{Limitations.}
\label{sec:supp-limitations}
Our comparisons rest on publicly available checkpoints trained with
heterogeneous data scales, compute budgets, and augmentation
recipes, confounded with the pretraining objective itself, so
family-level conclusions should be read with this caveat in mind. We
also measure representation structure rather than downstream
pose-estimation accuracy, and isolate rotation as the primary source
of pose variation; translating our findings into a complete pipeline,
or extending the analysis to uncropped scenes where translation and
scale also vary, is left to future work.

\paragraph{Future work.}
The cross-family complementarity we observe (main paper, Sec.~4.5)
motivates feature-fusion across pretraining objectives, and the
layer-depth trade-off we identify motivates adaptive layer selection
over a single fixed choice per deployment.

\end{document}